\documentclass[preprint]{elsarticle}
\usepackage{CJKutf8}

\usepackage{amsmath,amssymb,amsfonts}
\usepackage{tikz}
\usetikzlibrary{positioning, calc, decorations.pathreplacing}
\usepackage{pdflscape}
\usepackage{booktabs}
\usepackage{multirow}
\usepackage{subcaption}

\definecolor{casblue}{RGB}{8,62,140}
\definecolor{casred}{RGB}{203,65,84} 
\definecolor{casgreen}{RGB}{34,139,34} 
\definecolor{casyellow}{RGB}{218,165,32} 
\definecolor{caspurple}{RGB}{148,0,211} 
\mathchardef \mhyphen = "2D

\newcommand{\softmax}{\operatorname{softmax}}
\newcommand{\sigmiod}{\operatorname{sigmoid}}

\journal{Expert Systems with Applications}

\begin{document}

\begin{frontmatter}

\title{A Unified Framework of Medical Information Annotation and Extraction for Chinese Clinical Text}


\author[hwamei,institute]{Enwei Zhu}
\ead{zhuenwei@ucas.ac.cn}

\author[hwamei]{Qilin Sheng}
\ead{shengqilin0106@163.com}

\author[hwamei]{Huanwan Yang}
\ead{yanghuanw@163.com}

\author[hwamei,institute]{Jinpeng Li\corref{mycorrespondingauthor}}
\cortext[mycorrespondingauthor]{Corresponding author}
\ead{lijinpeng@ucas.ac.cn}

\address[hwamei]{HwaMei Hospital, University of Chinese Academy of Sciences, Ningbo 315010, Zhejiang Province, P. R. China}
\address[institute]{Ningbo Institute of Life and Health Industry, University of Chinese Academy of Sciences, Ningbo 315016, Zhejiang Province, P. R. China}

\begin{abstract}
Medical information extraction consists of a group of natural language processing (NLP) tasks, which collaboratively convert clinical text to pre-defined structured formats. This is a critical step to exploit electronic medical records (EMRs). Given the recent thriving NLP technologies, model implementation and performance seem no longer an obstacle, whereas the bottleneck locates on the whole engineering workflow and high-quality annotated corpora. This study presents an engineering framework consisting of three tasks, i.e., medical entity recognition, relation extraction and attribute extraction. Within this framework, the whole workflow is demonstrated from EMR data collection through model performance evaluation. Our annotation scheme is comprehensive and compatible between tasks; the resulting corpus, which was manually annotated by experienced physicians, is of large scale and high quality. Built upon this corpus, the medical information extraction system show performance that approaches human annotation. The annotation scheme, (a subset of) the annotated corpus, and the code are all publicly released, to facilitate further research. 
\end{abstract}


\end{frontmatter}


\section{Introduction} \label{sec:intro}
It has been widely recognized that electronic medical record (EMR) is a firsthand, evidential, valuable source of clinical experience and medical knowledge, which is useful for either scientific investigations \cite{casillas2016learning,sung2020emr} or industrial applications \cite{liang2019evaluation}. However, EMRs are in general underutilized in practice, because of the issue of under-structuralization \cite{ravi2017deep,pramanik2020healthcare}. More specifically, the clinic-related contents are written and stored in free text, which is flexible and potentially ambiguous; such data structure creates obstacles for medical information retrieval by clinic-related queries, e.g., ``what are the potential side effects of aspirin?'' Medical information extraction aims to fill this gap, by converting the free-text contents in EMRs to pre-defined structured formats. 

However, it still remains a great challenge to establish a mature medical information extraction system. Although it seems no longer difficult to implement a high-performance information extraction algorithm, thanks to the vastly developing natural language processing (NLP) community \cite{lecun2015deep,hirschberg2015advances}; creating a high-quality annotated corpus to train and evaluate the NLP models, becomes the bottleneck. The essential engineering principles and details for medical information annotation are scattered in the literature \cite{uzuner2011i2b2,roberts2009building,albright2013towards,he2017building}, and limited studies have demonstrated how to connect different stages, i.e., design, annotation, modeling and evaluation, in the whole workflow. 

In addition, the contents and writing styles of EMRs are significantly heterogeneous across hospitals of different sizes, levels and specialties. Hence, an information extraction system well optimized for one hospital may encounter severe incompatibility when transferring to other hospitals, and thus requires heavy customized engineering, or worse, developing from scratch; this is quite different from the AI systems for medical images (e.g., X-ray or CT scan). This further emphasizes the importance of a standard operation protocol (SOP) with detailed demonstration, especially to the practitioners new to this field.


This study proposes a unified framework of medical information annotation and extraction, consisting of three tasks: entity recognition, relation extraction and attribute extraction. We describe the whole workflow of corpus construction, from the collection and cleaning of EMRs, the philosophy and details of annotation scheme, the annotation tools and practice, to the resulting corpus. The EMRs come from HwaMei Hospital of University of Chinese Academy of Sciences, a general hospital in Ningbo, China. The corpus was manually annotated by experienced physicians, and includes 1,200 full medical records (referred as HwaMei-1200 henceforth); to the best of our knowledge, this scale is larger than all those in previous studies. The inter-annotator agreements (IAAs) reach 94.53\%, 73.73\% and 91.98\% $F_1$ scores for entity, relation and attribute annotations, respectively. 

\begin{figure}
    \centering
    \includegraphics[width=\textwidth]{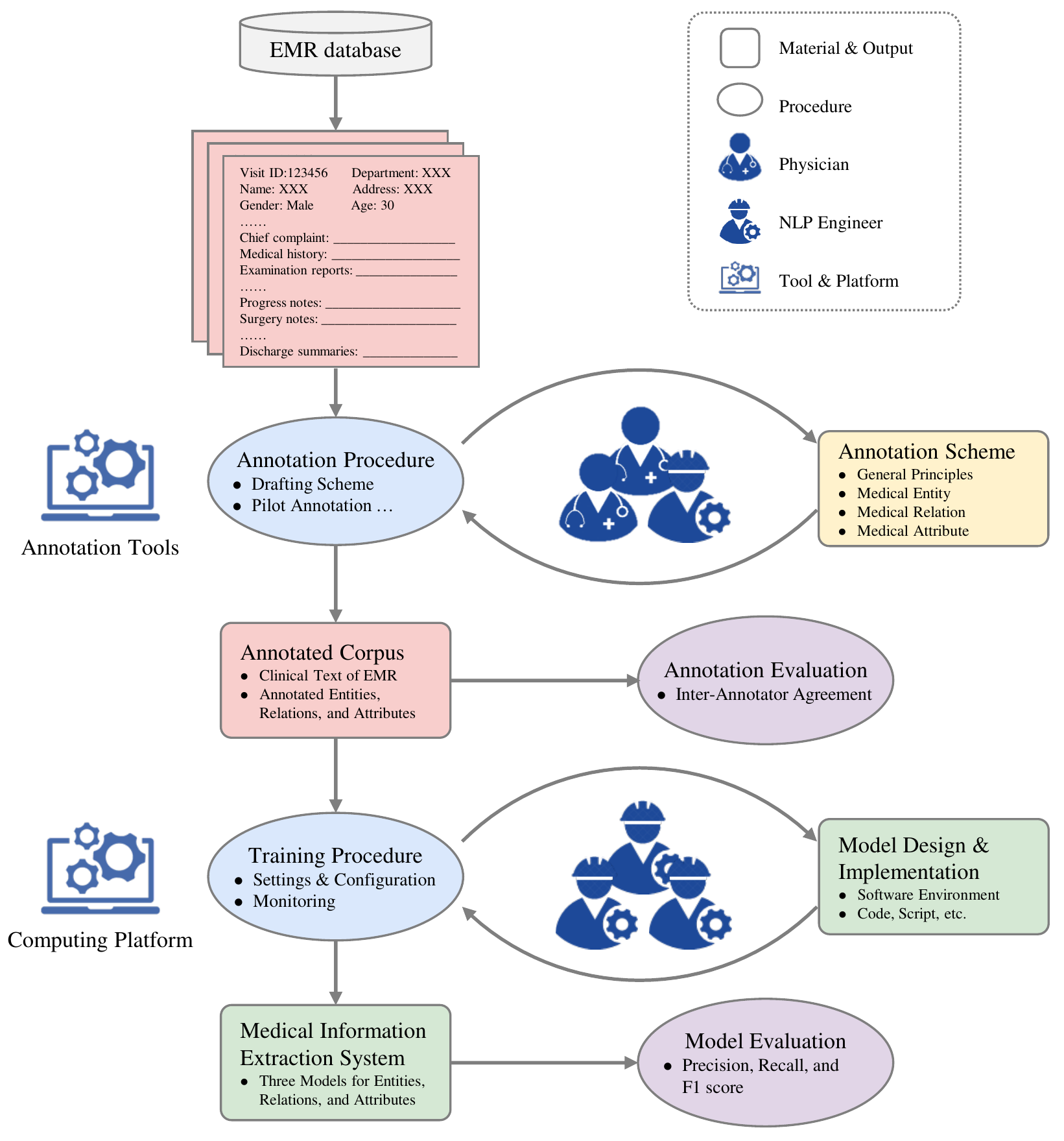}
    \caption{The workflow for a medical information extraction system}
    \label{fig:workflow}
\end{figure}

We then demonstrate how to utilize the corpus to develop a medical information extraction system, based on classic neural NLP models, e.g., LSTM \cite{hochreiter1997long} and BERT \cite{devlin2019bert}. We perform experiments on a random subset of 500 medical records, and the best $F_1$ scores of entity, relation and attribute extractions reach 92.23\%, 61.46\% and 87.51\%, respectively; additionally, training with more data will consistently and progressively improve the performance across all three tasks. Figure \ref{fig:workflow} presents the whole workflow described above. To facilitate further research, we release the detailed version of our annotation scheme, the 500 annotated medical records (referred as HwaMei-500 henceforth), and the code.\footnote{The code and annotation scheme are available at \underline{https://github.com/syuoni/eznlp}. The corpus is available upon request and signing a data use agreement. It has been manually de-identified.} 

This study contributes to the literature from two folds:
\begin{itemize}
    \item For practitioners, we demonstrate a feasible workflow from raw EMRs to a medical information extraction system. Especially in case of Chinese EMRs, our released data and code may serve as a decent starting point. Given the annotation scheme available, it is easy to extend the corpus with their own EMRs if higher model performance or customized engineering is required. We also present how the model performance will increase with more training data. We believe that these will provide useful and reliable guidance for practice. 
    \item For researchers, we provide a benchmark to evaluate their information extraction algorithms. The corpus is from real EMRs, and carefully annotated by experienced physicians, covering a range of medical departments and EMR sections. Hence, the corresponding evaluation results are indicative and representative. In addition, this benchmark naturally supports multi-task modeling. 
\end{itemize}

This article proceeds as follows. The next section reviews the literature on clinical text corpora for medical information extraction. Section \ref{sec:construction} describes the designs and details in the workflow of corpus construction, especially for the annotation scheme. Section \ref{sec:extraction} demonstrates how to use the corpus to develop models for medical information extraction, and reports the experimental results. Section \ref{sec:conclusion} concludes.

\section{Literature Review} \label{sec:literature}
Large-scale and high-quality annotated datasets are one of the most important foundations for the success of machine learning modeling, especially for deep learning models \cite{lecun2015deep}. Current state-of-the-art (SOTA) NLP models are highly integrated with deep learning techniques \cite{hirschberg2015advances}. Hence, medical information extraction, as a group of NLP tasks, also relies on large-scale and high-quality annotated clinical corpora. Extensive public corpora have been released and significantly facilitated related algorithm research, such as CoNLL 2003 \cite{sang2003introduction} for named entity recognition, and ACE 2004/2005 for relation extraction. In addition, a number of corpora are available in the biomedical domain, such as GENIA \cite{kim2003genia}, NCBI \cite{dougan2014ncbi} and BC5CDR \cite{li2016biocreative}, of which the data sources are typically PubMed articles. However, clinical text has significantly different syntactic and semantic patterns from either common text or biological text, so the clinical NLP systems rely on separate clinical text corpora. 

The clinical corpora in English have been relatively extensive and easily available. The famous i2b2 (Informatics for Integrating Biology to the Bedside) project organized a series of shared task challenges and publicly released corresponding corpora, which significantly facilitated medical information extraction for clinical narratives. The i2b2 shared tasks include obesity status recognition \cite{uzuner2009recognizing}, extracting medication-related information \cite{uzuner2010extracting}, medical concepts, assertions and relations \cite{uzuner2011i2b2}, and temporal relation extraction \cite{sun2013evaluating}. The 2010 i2b2 challenge \cite{uzuner2011i2b2} is the first to propose a relatively comprehensive framework for medical information extraction. Specifically, the framework is characterized by three entity types (medical problem, test and treatment), six attribute types (present, absent, possible, conditional, hypothetical and associated with others), and 11 relation types. Many related studies followed up and extended this framework. 

Meystre and Haug constructed a clinical text corpus consisting of 160 documents, which are annotated with medical problems and corresponding assertions of being present or absent \cite{meystre2006natural}. Ogren \emph{et al.} proposed a similar corpus of 160 clinical notes, and extended the assertions to six types (current, history of and family history of, confirmed, possible and negated) \cite{ogren2008constructing}. Roberts \emph{et al.} proposed the CLEF (CLinical E-Science Framework) corpus, which includes 150 annotated documents \cite{roberts2009building}. The CLEF annotation scheme proposes six entity types (condition, intervention, investigation, result, drug or device, and locus), and eight relation types. The scheme does not include attribute annotation, but is able to formulate a negation attribute by attaching a ``modifier'' relation between the negation word and corresponding entity. Albright \emph{et al.} and Savkov \emph{et al.} aimed to construct clinical corpora covering both syntactic and semantic annotations \cite{albright2013towards, savkov2016annotating}. In addition, some recent studies began to focus on clinical events and their temporal relations \cite{sun2013evaluating}. 

Studies emerge on medical information extraction for Chinese clinical data in recent years. CCKS (China Conference on Knowledge Graph and Semantic Computing) released a shared task of named entity recognition for Chinese EMRs. The corpus includes six entity types, and its 2019 version consists of about 1,400 annotated documents. In addition, He \emph{et al.} designed a comprehensive syntactic and semantic annotation scheme covering four entity types, 15 relation types and seven assertion types; the corpus consists of 992 documents of discharge summaries and progress notes \cite{he2017building}. Gao \emph{et al.} employed 225 admission records for medical entity annotation \cite{gao2019constructing}. Guan \emph{et al.} proposed the CMeIE dataset, using medical textbooks and clinical notes as the data source; the annotation scheme consists of 11 entity types and 44 relation types \cite{guan2020cmeie}. Lee and Lu collected textual data from health-related news, magazines and question/answer forums, and annotated the medical entities of ten types \cite{lee2021multiple}. 

Table \ref{tab:corpara-review} summarizes the clinical text corpora proposed in literature. Typically, a comprehensive information extraction framework comprises three medical information elements, i.e., entity, relation and attribute. Medical entity is the most basic information element. Early studies focused on a single entity type such as medical problem \cite{meystre2006natural, ogren2008constructing} or medication \cite{uzuner2010extracting}, whereas a relatively comprehensive scheme would include at least disease, symptom, test, treatment and drug \cite{roberts2009building, uzuner2011i2b2}. Medical relation types are defined between each pair of entity types. This typically yields a large number of relation types \cite{campillos2018french, guan2020cmeie}, which may easily cause ambiguity between different types and thus confuse the annotators. Medical attribute is also called assertion, which indicates some special statuses of an entity. The most basic attribute types are negation and uncertainty \cite{uzuner2011i2b2}. In general, these annotation schemes are either incomprehensive or short of consistency; specifically, the entity, relation and attribute types are typically separately defined, with limited illustration of the compatibility between them. Their corpus scales are relatively small, which may not be sufficient to build a reliable information extraction system. Further, they hardly describe how the design, annotation, modeling and evaluation stages are connected as a whole engineering project. 

We propose a unified framework of clinical-text-oriented information annotation and extraction. The annotation scheme includes nine entity types (18 subtypes), ten relation types and ten attribute types, as well as the compatibility defined between them. The resulting corpus (HwaMei-1200) consists of 1,200 full inpatient records (or 18,039 broken-down documents), which significantly surpasses the scales of most previous corpora. A subset of 500 records (HwaMei-500) is randomly selected for public release. We further demonstrate how to utilize the corpus for modeling. These form a comprehensive operation protocol towards a medical information system. 

\begin{landscape}
\begin{table}
    \centering
    \caption{Clinical text corpora for medical information extraction}
    \begin{tabular}{llp{150pt}p{95pt}ccc}
        \toprule
        Corpus & Language & Material & Size & \#Entities & \#Relations & \#Attributes \\
        \midrule
        Meystre and Haug \cite{meystre2006natural}         & English & Clinical notes & 160 documents & 1 & -- & 2 \\
        Ogren \emph{et al.} \cite{ogren2008constructing}   & English & Outpatient notes, discharge summaries, inpatient notes & 160 documents & 1 & -- & 6 \\
        Roberts \emph{et al.} \cite{roberts2009building}   & English & Clinical narratives, histopathology reports, imaging reports & 150 documents & 6 & 8 & -- \\
        Uzuner \emph{et al.} \cite{uzuner2010extracting}   & English & Discharge summaries & 1,243 documents & 7 & -- & --\\
        Uzuner \emph{et al.} \cite{uzuner2011i2b2}         & English & Discharge summaries, progress reports & 871 documents & 3 & 11 & 6 \\
        Albright \emph{et al.} \cite{albright2013towards}  & English & Clinical notes, pathology notes & 13,091 sentences & 16 & -- & --\\
        Savkov \emph{et al.} \cite{savkov2016annotating}   & English & Primary care notes & 750 documents & 4 & -- & -- \\
        CCKS 2019                                     & Chinese & Clinical notes & 1,400 documents & 6 & -- & -- \\
        He \emph{et al.} \cite{he2017building}        & Chinese & Discharge summaries, progress notes & 992 documents & 4 & 15 & 7 \\
        Gao \emph{et al.} \cite{gao2019constructing}  & Chinese & Admission records & 225 documents & 9 & -- & -- \\
        Guan \emph{et al.} \cite{guan2020cmeie}       & Chinese & Medical textbooks, clinical notes & 28,008 sentences & 11 & 44 & -- \\
        Lee and Lu \cite{lee2021multiple}             & Chinese & Health-related news, magazines and question/answer forums & 30,692 sentences & 10 & -- & -- \\
        \midrule
        HwaMei-1200 (ours)                            & Chinese & Inpatient records & 1,200 full records \par (18,039 documents, \par or 130,945 sentences) & 18 & 10 & 10 \\
        HwaMei-500 (ours)                             & Chinese & Inpatient records & 500 full records & 18 & 10 & 10 \\
        \bottomrule
    \end{tabular}
    \label{tab:corpara-review}
\end{table}
\end{landscape}

\section{Corpus Construction} \label{sec:construction}
\subsection{Data Source}
Our corpus material came from HwaMei Hospital of University of Chinese Academy of Sciences, a general hospital in Ningbo, China. Its hospital information system (HIS) has been in service for about ten years, and a large amount of EMRs have been accumulated. We obtained a 785 GB backup of 109.9 thousand inpatient records between September 2019 and December 2020 (16 months). After data cleaning and parsing, there remained 77.9 thousand valid records. Each inpatient record is indexed by a unique visit ID, and contains many structured fields like the patient's name, demographics (age, gender, etc.) and administrative information (medical department, bed number, admission and discharge dates, etc.). However, clinic-related fields are semi-structured; that is, they are written and stored as free text in specific sections, such as chief complaint, medical history, examination reports, progress notes, surgery notes and discharge summaries. 

As suggested by previous studies, a corpus should be \emph{representative} and \emph{balanced} \cite{campillos2018french}. Hence, a good clinical corpus should cover the variety of language use in clinical text, and different syntactic and semantic patterns should share approximately equal weights in the distribution. However, the natural distribution of clinical text is unbalanced. In particular, a few common diseases take a very high proportion of the medical record population, and the medical records of a same disease are virtually identical in the treatment progress, resulting in very similar language use in clinical text. Therefore, we selected our annotation samples via a stratified sampling procedure assisted by manual check. Specifically, we firstly chose 30 representative surgery and non-surgery departments, and randomly sampled hundreds of medical records from each department. Then, we manually checked the samples for each department; if a specific disease appeared too frequently, we would randomly select a small number of such samples, and discarded the remaining ones. 

\begin{figure}
    \centering
    \includegraphics[width=\columnwidth]{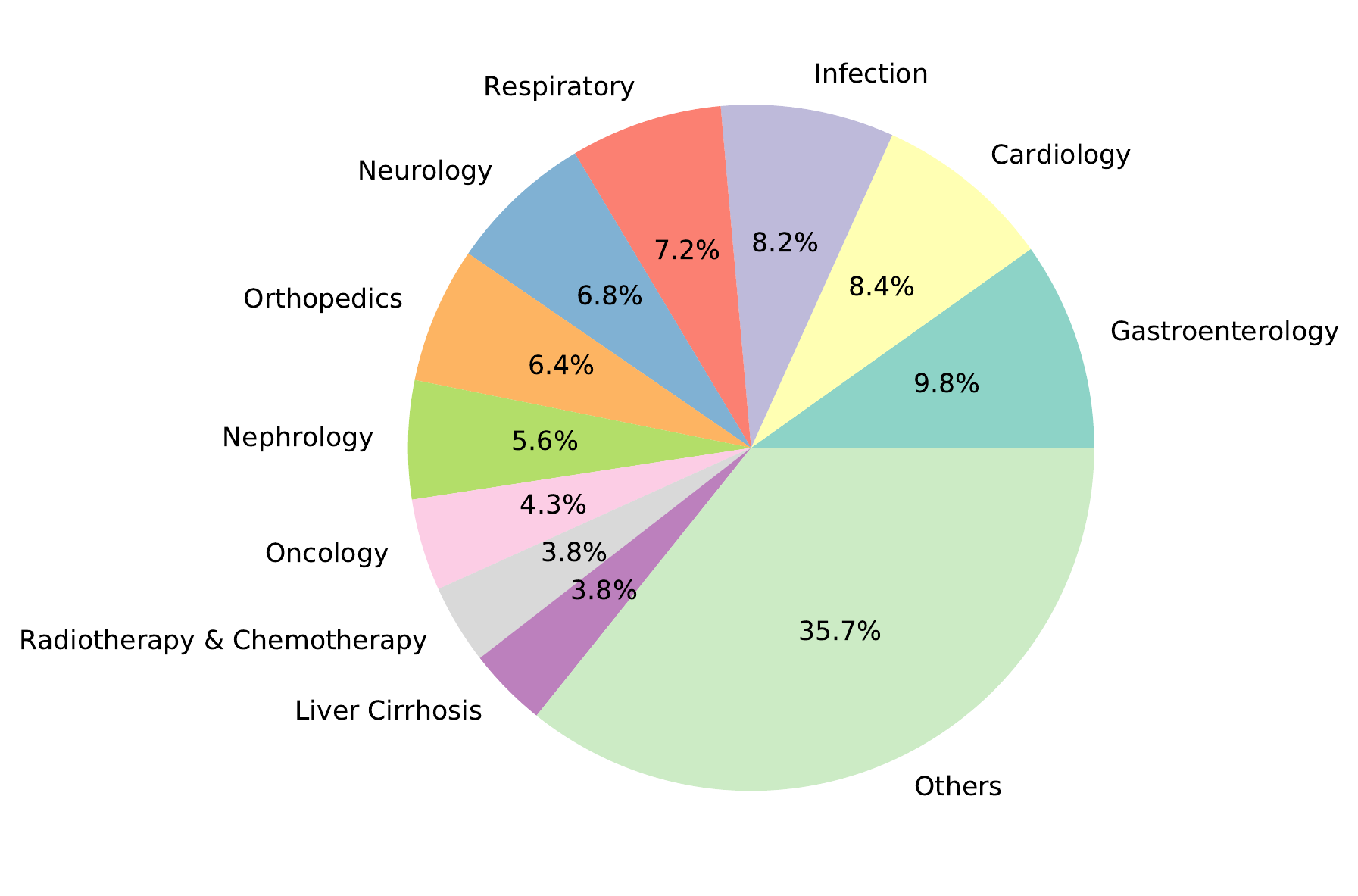}
    \caption{Medical department distribution of annotated corpus}
    \label{fig:depart-destr}
\end{figure}

The final corpus consists of 1,200 medical records, which cover the common diseases in 30 representative medical departments. Figure \ref{fig:depart-destr} presents the department distribution, in which the gastroenterology, cardiology, infection and respiratory departments represent relatively high proportions. Note that each record is complete in all fields and sections. We regard the text in each section as a document; then the corpus includes 18,039 documents, or can be further broken down into 130,945 sentences. 

\subsection{Annotation Scheme}
We proposed a unified medical information annotation scheme for Chinese clinical text, which was based on comprehensive reviews of previous studies \cite{uzuner2011i2b2, roberts2009building, he2017building, campillos2018french} and multiple-round consultations with experienced medical experts. This subsection provides a brief description, whereas the detailed annotation guidelines and extensive examples are available as supplementary material. 

\subsubsection{General Principles}
In general, we followed these annotation principles: 

\paragraph{Collectively exhaustive} The definitions of the medical entities, relations and attributes should be as comprehensive as possible, so that the extracted elements are capable of describing and representing the medical information in clinical text. Ideally, all the medical information in the original clinical text can be retrieved via the extracted medical elements. 

\paragraph{Mutually exclusive} The definitions of medical entities, relations and attributes should be as clear as possible at the boundaries between different types. It should be avoided that an entity mention (or an entity pair) can be legally assigned to multiple entity (or relation) types, which will confuse the annotators and lead to inconsistent annotation results. 

\paragraph{Semantics are contextualized} The clinical text should be interpreted along with its contextual information. Even a same word or phrase may express completely different semantics and thus require different annotations, given the contexts are different. In particular, the annotator should not mechanically map local words and phrases to fixed annotation results, just like looking up a fixed table. For example, any abbreviation or descriptive phrase should be annotated as the corresponding medical element, provided that the context is clear \cite{uzuner2011i2b2}. 

\paragraph{Literalness is prior to external knowledge} The annotations should firstly accord with the text literalness, and secondly the annotator's medical knowledge background. As an extreme (and infrequent) case, the text may be vague or even have some mistakes, and thus conflict with the annotator's medical knowledge; in this case, the annotator should follow the literalness. Otherwise, if the text implicitly mentions a medical element, or in other words, it can be inferred according to the annotator's medical knowledge, it should be annotated only when the knowledge is a widely recognized consensus (e.g., the normal body temperature range) \cite{roberts2009building}. 

These principles are somewhat philosophy of designing a scheme. In practice, the annotators reported that emphasizing these principles could help them to understand how the guidelines were designed, and thus improve their annotation efficiency and quality.

\subsubsection{Medical Entity}
Given an input text as a sequence of tokens $x = x_1,x_2,\dots,x_T$ of length $T$, the objective output of entity recognition is a set $y^{Ent}$, of which each element is a triplet of entity type, start position, and end position: 
\begin{align} \label{eq:entity}
    y^{Ent} = \{ (type^{Ent}_i, start_i, end_i) \mid type^{Ent}_i \in S^{Ent}, \nonumber \\ 0 \leq start_i, end_i < T, i=1,2,\dots,N \},
\end{align}
where $S^{Ent}$ is the set of entity types, $N$ is the number of entities in the text. Correspondingly, the annotation of an entity should specify its type, start and end positions. 

We formulated nine entity types, and an entity type may consist of multiple subtypes, which further characterize the entity natures. Their definitions and a few simple examples are listed as follow: 
\begin{CJK*}{UTF8}{gbsn}
\begin{enumerate}
    \item \textbf{Disease}: A condition or status which impairs normal functioning, typically manifested in distinguishing symptoms and signs
    \begin{enumerate}
        \item \textbf{Disease or Syndrome}: Diseases mainly developed by internal causes; e.g., 心脏病~(heart disease), 糖尿病~(diabetes)
        \item \textbf{Injury or Poisoning}: Diseases mainly developed by external causes; e.g., 头部外伤~(head trauma)
        \item \textbf{Organ Damage}: Descriptions of organ damages, dysfunctions, disturbances, insufficiencies, etc., with limited implication of the causes; e.g., 肾损害~(kidney damage), 肝功能不全~(liver insufficiency)
    \end{enumerate}
    \item \textbf{Symptom}: A phenomenon or manifest which is typically caused by, and in turn indicates the existence of diseases
    \begin{enumerate}
        \item \textbf{Self-Reported Abnormality}: A subjective discomfort reported by the patient; e.g., 乏力~(fatigue), 胃痛~(stomach ache)
        \item \textbf{Abnormal Test Result}: An abnormal evidence found by tests; e.g., 腹腔积液~(fluid in abdominal cavity), 血压过高~(high blood pressure)
    \end{enumerate}
    \item \textbf{Test}: A process for searching clues of diseases or its causes, including physical examination, laboratory and imaging tests, etc.
    \begin{enumerate}
        \item \textbf{Test Process}: A test service, which may refer to one test item, or a group of items; e.g., 血常规~(blood routine), X光检查~(X-ray examination)
        \item \textbf{Test Result}: A test item with the result; e.g., 脉搏52次/分~(pulse 52 beats/min), 血压125/86mmHg~(blood pressure 125/86mmHg)
    \end{enumerate}
    \item \textbf{Treatment}: A process for curing diseases or relieving symptoms
    \begin{enumerate}
        \item \textbf{Treatment}: A complete treatment process, such as surgeries, chemotherapy and radiotherapy; e.g., 化疗~(chemotherapy), 切开排脓术~(incision and drainage)
        \item \textbf{Operation}: A step of treatment, which does not form a complete treatment process; e.g., 局麻~(local anesthesia), 止血~(stop bleeding)
        \item \textbf{Prevention}: A treatment process aimed to prevent potential diseases or symptoms in the future; e.g., 抗凝血~(anticoagulant), 抗休克~(anti-shock)
        \item \textbf{Care}: A nursing care process; e.g., I级护理~(level I care), 普食~(general food)
    \end{enumerate}
    \item \textbf{Drug}: A medicine for curing diseases or relieving symptoms
    \begin{enumerate}
        \item \textbf{Drug}: The drug name with its specification; e.g., 止疼片~(painkiller), 1\%利多卡因~(1\% lidocaine)
        \item \textbf{Drug Dose}: The drug quantity (including the number and unit); e.g., 2片一次，一日3次~(2 tablets once, 3 times a day)
    \end{enumerate}
    \item \textbf{Body}: A body part or matter
    \begin{enumerate}
        \item \textbf{Body Part}: An organ, tissue or location of a human body; e.g., 上肺尖~(upper lung apex)
        \item \textbf{Body Matter}: Body fluid, secretion, urine or excrement; e.g., 胃内容物~(stomach contents)
    \end{enumerate}
    \item \textbf{Personal History}: Patient's history activities related to medical problems; e.g., 吸烟史~(smoking history)
    \item \textbf{Equipment}: Medical equipment, instruments and devices; e.g., 电刀~(electric knife)
    \item \textbf{Department}: The name of a medical department; e.g., 心内科~(cardiology)
\end{enumerate}
\end{CJK*}

A Test Result entity is assumed to be \emph{normal} or \emph{negative} (i.e., within the range of a healthy status), whereas an Abnormal Test Result entity should be \emph{abnormal} or \emph{positive} (i.e., deviated from a healthy status). In other words, these two types are the negated versions of each other (see Subsection \ref{subsubsec:attr}). It may appear not so straightforward that they separately belong to two different super types. The underlying logic is that, the Abnormal Test Result is very close to the concept of ``signs'' in clinical practice, which is a manifest of diseases, just like Self-Reported Symptom; hence, they typically attract much more physician's attention than a negative Test Result. 

We employed a \emph{flat} scheme for entity annotation, which means that no nested or overlapping entities are allowed. This was a trade-off result; otherwise, there would be too many trivial and redundant entities. For example, it is a common case that a Disease or Symptom entity includes a Body Part entity in its name. The Disease or Symptom entity is a whole concept, which has already indicated the corresponding body part, so explicitly annotating the nested Body Part entity provides limited incremental information. 

To this end, we should design which entities should be preferentially annotated in potential clashing cases. In general, the former five entity types (Disease, Symptom, Test, Treatment and Drug) were regarded to be more important than the later four types (Body, Personal History, Equipment and Department); hence, the former five types are of relatively high priority in annotation. For example, in text ``the patient has heart disease \dots'', a Body Part entity is inside a Disease entity, then the prior Disease entity ``heart disease'' should be annotated whereas the Body Part entity ``heart'' should be left unannotated. 

It is common that a medical entity appears as an abbreviation in clinical text. These abbreviations should be interpreted and annotated as the corresponding original entities. For example, in text ``the patient should check the kidney function and blood sugar \dots'', the span ``blood sugar'' does refer to a medical test based on the context, although the span itself may not necessarily refer to it. In this case, we would still treat the text span as a valid mention of the entity.

\subsubsection{Medical Relation}
Given tokens $x$ and entities $y^{Ent}$, the objective output of relation extraction is a set $y^{Rel}$, of which each element is a triplet of relation type, head entity and tail entity: 
\begin{align} \label{eq:relation}
    y^{Rel} = \{ (type^{Rel}_j, head_j, tail_j) \mid type^{Rel}_j \in S^{Rel}, \nonumber \\ head_j, tail_j \in y^{Ent}, j=1,2,\dots,M \},
\end{align}
where $S^{Rel}$ is the set of relation types, $M$ is the number of relations between the entities. Correspondingly, the annotation of a relation should specify its relation type, head and tail entities. 

Previous studies typically enumerated every pair of entity types and defined multiple relation types between each pair \cite{uzuner2011i2b2, he2017building, campillos2018french}. This approach may easily lead to either omissions or redundancies of the relation type coverage, especially when the entity types are defined of a fine granularity. An entity of a specific type may play different roles 
relative to another entity, which results in many possible relation types. Based on multiple-round in-depth discussion with medical experts, we formulated medical relations between three essential ``entity roles'', namely ``Status'', ``Information'' and ``Intervention''. 
\begin{enumerate}
    \item \textbf{Status}: The patient's essential, underlying health conditions, which may cause certain relatively non-essential appearances. A Status role is typically played by a Disease, Symptom or Personal History entity. 
    \item \textbf{Information}: The patient's non-essential appearances, which is typically caused by relatively essential underlying health conditions. It can be further divided into three sub roles. 
    \begin{enumerate}
        \item \textbf{Positive Information}: An abnormal appearance, typically played by a Disease, Symptom (including Self-Reported Abnormality and Abnormal Test Result) or Test Process entity. 
        \item \textbf{Negative Information}: A normal appearance, typically played by a Test Result or Test Process entity.
        \item \textbf{Unknown Information}: An appearance unknown whether it is normal or not, typically played by a Test Process entity.  
    \end{enumerate}
    \item \textbf{Intervention}: The physician's measures aimed at improving the patient's health conditions. An Intervention role is typically played by a Treatment or Drug entity. 
\end{enumerate}

\begin{figure}
    \centering
    \begin{tikzpicture}[scale=1.0, transform shape, 
                        recnode/.style = {rectangle, thick, rounded corners=2pt, minimum size=15pt, align=left}, 
                        stedge/.style  = {-stealth, draw=gray!70, thick, fill=gray!70}]
        \node[recnode, draw=casred!70, fill=casred!20]     (status) at (-80pt, 0pt) {Status \\ 
                                                                                     \scriptsize{\textendash \enspace Disease} \\ 
                                                                                     \scriptsize{\textendash \enspace Symptom} \\
                                                                                     \scriptsize{\textendash \enspace Personal History}};
        \node[recnode, draw=casblue!70, fill=casblue!20]   (info)   at (80pt, 0pt) {Information \\ 
                                                                                    \scriptsize{\textendash \enspace Disease (+)} \\ 
                                                                                    \scriptsize{\textendash \enspace Symptom (+)} \\ 
                                                                                    \scriptsize{\textendash \enspace Test Process (+/\textminus/*)} \\
                                                                                    \scriptsize{\textendash \enspace Test Result (\textminus)}};
        \node[recnode, draw=casgreen!70, fill=casgreen!20] (interv) at (0pt, 80*1.732 pt) {Intervention \\ 
                                                                                           \scriptsize{\textendash \enspace Treatment} \\ 
                                                                                           \scriptsize{\textendash \enspace Drug}};
        \path[stedge] (status) edge[bend left=15pt] (info);
        \node[yshift=22.5pt, align=center, font=\scriptsize] at (0pt, 0pt) {Cause (+) \\ Require (*)};
        \path[stedge] (info) edge[bend left=15pt] (status); 
        \node[yshift=-22.5pt, align=center, font=\scriptsize] at (0pt, 0pt) {Suggest (+/\textminus) \\ Exclude (+/\textminus)};
        \path[stedge] (status) edge[bend left=15pt] (interv);
        \node[xshift=-25pt, yshift=10pt, align=center, font=\scriptsize] at (-40pt, 40*1.732 pt) {Require};
        \path[stedge] (interv) edge[bend left=15pt] (status);
        \node[xshift=25pt, yshift=-5pt, align=center, font=\scriptsize] at (-40pt, 40*1.732 pt) {Cause \\ Modify};
        \path[stedge] (info) edge[bend left=15pt] (interv);
        \node[xshift=-25pt, yshift=-20pt, align=center, font=\scriptsize] at (40pt, 40*1.732 pt) {Permit (+/\textminus) \\ Contra (+/\textminus)};
        \path[stedge] (interv) edge[bend left=15pt] (info);
        \node[xshift=25pt, yshift=10pt, align=center, font=\scriptsize] at (40pt, 40*1.732 pt) {Require (*)};
    \end{tikzpicture}
    \caption{Medical relations between entities as Status, Information and Intervention.
     The positive/negative/star sign marked on the right of an entity type under the Information role indicates that the entity type can play Positive/Negative/Unknown Information. The positive/negative/star sign marked on the right of a relation type indicates that the relation applies to Positive/Negative/Unknown information. If an entity type is listed, the corresponding relations apply to all its subtypes.}
    \label{fig:relations}
\end{figure}
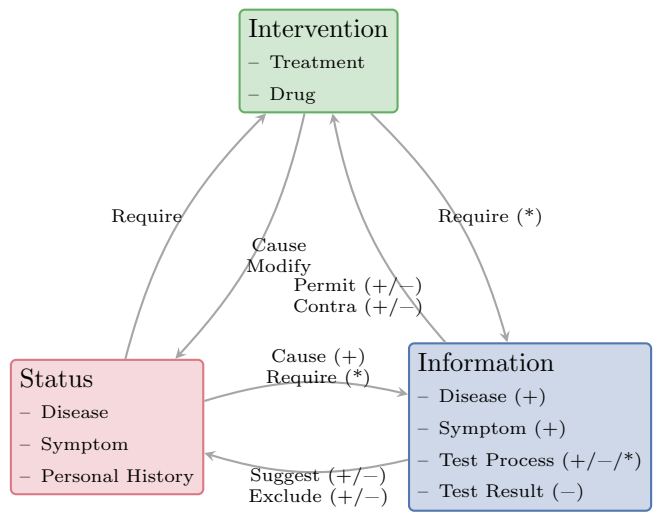

Figure \ref{fig:relations} presents the potential medical relations between the three roles. We argue that most essential relation types have been covered in this framework. Their definitions and examples are listed as follow: 
\begin{CJK*}{UTF8}{gbsn}
\begin{enumerate}
    \item \textbf{Status--Cause--Information}: The Status entity is one of the potential reasons or triggers of the Information entity, where the Information is supposed to be positive; e.g., 高血压容易引发头晕~(High blood can easily cause dizziness)
    \item \textbf{Status--Require--Information}: The Status entity requires the Information entity to confirm or exclude, where the Information is supposed to be unknown. Accordingly, the context should not explicitly mention the result (i.e., confirming or excluding the Status entity), since the Information entity is still unknown; otherwise, the relation should be annotated as Information--Suggest--Status or Information--Exclude--Status described below. e.g., 消化性溃疡可通过查胃镜助诊~(Peptic ulcer can be diagnosed with gastroscopy)
    \item \textbf{Information--Suggest--Status}: The Information entity suggests the existence of the Status entity, where the Information could be either positive or negative; e.g., 患者患有黄疸型肝炎，考虑药物性肝损可能性大~(The patient has jaundice hepatitis, and the possibility of drug-induced liver damage should be considered)
    \item \textbf{Information--Exclude--Status}: The Information entity excludes the possibility of the Status entity, where the Information could be either positive or negative; e.g., 消化性溃疡：常有……症状，该患者已作胃镜排除~(Peptic ulcer: patients often have \dots symptoms, which has been excluded by gastroscopy)
    \item \textbf{Status--Require--Intervention}: The Status entity requires the Intervention entity to modify, such as a disease requiring treatment. The context should not explicitly mention the result of intervention; otherwise, the relation should be annotated as Intervention--Modify--Status described below; e.g., 患高血压，平素服用缬沙坦~(The patient has hypertension and usually takes valsartan)
    \item \textbf{Intervention--Modify--Status}: The Intervention entity modifies the Status entity. According to the result, this relation type can be further divided into Intervention--Improve--Status and Intervention--Worsen--Status; e.g., 患者被诊断为三叉神经痛，在三叉神经神经射频热凝术后，疼痛好转~(The patient was diagnosed with trigeminal neuralgia, and it got alleviated after radiofrequency thermocoagulation of the trigeminal nerve)
    \item \textbf{Intervention--Cause--Status}: The Intervention entity causes the Status entity, which is typically a complication; e.g., 患者化疗后出现感染性发热~(The patient developed infectious fever after chemotherapy)
    \item \textbf{Intervention--Require--Information}: The Intervention entity requires the Information entity to confirm its feasibility, where the Information is supposed to be unknown. Accordingly, the context should not explicitly mention the confirming result, since the Information entity is still unknown; otherwise, the relation should be annotated as Information--Permit--Intervention or Information--Contraindicate--Intervention described below. e.g., 损伤处查CT或MRI，排除手术禁忌~(Check CT or MRI to exclude surgical contraindications)
    \item \textbf{Information--Permit--Intervention}: The Information entity confirms the feasibility of the Intervention entity, where the Information could be either positive or negative; e.g., 如无明显低血糖，必要时可使用瑞格列奈片~(If there is no obvious hypoglycemia, repaglinide tablets can be used if necessary) 
    \item \textbf{Information--Contraindicate--Intervention}: The Information entity contraindicates the application of the Intervention entity, where the Information could be either positive or negative; e.g., 内固定取出后有再骨折风险，暂不考虑手术~(After the internal fixation is removed, there is a risk of re-fracture, so surgery is not considered for the time being) 
\end{enumerate}
\end{CJK*}

Table \ref{tab:rel-applicability} presents the ten medical relation types defined in our scheme, and their applicability to entity types. Note that if we define relation types between each pair of entity types, it will result in many semantically similar relation types. For example, if we specify Status--Cause--Information to all applicable entity types, there would be six relation types; if we specify it further to all sub types, there would be 30 relation types. Apparently, this would be too miscellaneous, easily resulting in omissions of the scheme and confusions to the annotators. Based on the abstraction of entity roles, we can formulate the relation types in a less error-prone and more elegant way. As presented, the resulted relation annotation scheme appears relatively simple, clear but comprehensive. 

\begin{landscape}
\begin{table}
    \caption{Medical relation types and applicable entity types}
    \centering
    \begin{tabular}{lll}
        \toprule
        Relation type                      & Head entity types                  & Tail entity types \\
        \midrule
        Status--Cause--Information         & Disease, Symptom, Personal History & Disease, Symptom \\
        Status--Require--Information       & Disease, Symptom                   & Test Process \\
        Information--Suggest--Status       & Disease, Symptom, Test Process, Test Result & Disease, Symptom \\
        Information--Exclude--Status       & Disease, Symptom, Test Process, Test Result & Disease, Symptom \\
        \midrule
        Status--Require--Intervention      & Disease, Symptom                   & Treatment, Drug \\
        Intervention--Modify--Status       & Treatment, Drug                    & Disease, Symptom \\
        Intervention--Cause--Status        & Treatment, Drug                    & Disease, Symptom \\
        \midrule
        Intervention--Require--Information & Treatment, Drug                    & Test Process \\
        Information--Permit--Intervention  & Disease, Symptom, Test Process, Test Result & Treatment, Drug \\
        Information--Contra--Intervention  & Disease, Symptom, Test Process, Test Result & Treatment, Drug \\
        \bottomrule
    \end{tabular}
    \label{tab:rel-applicability}
\end{table}
\end{landscape}

Entity relations are high-level semantics, so ideally, the confident annotation of a medical relation relies on both the correct text expression and the consistency with external knowledge. However, as reported by the annotators, the real clinical text contains a non-negligible quantity of typos, grammar mistakes and content errors, which may make it difficult to interpret the relations or cause conflicts between text meanings and external knowledge. As previously highlighted, the annotations should respect the text literalness more preferentially than the external knowledge. Hence, as long as the text clearly states a medical relation, the annotation should follow it; otherwise, the annotator can try to infer implicit relations from the text with the help of her medical knowledge. 

In our scheme, a Disease or Symptom entity may play a Status role or an Information role in a medical relation, depending on the contexts. Hence, given two entities, the annotator should first choose a role for each entity and then label their relation. In the case of two Disease (or Symptom) entities appearing in a relation, the annotator has to decide which is the relative essence and thus plays the Status role, and which is the relative appearance and thus plays the Information role. 

The relation annotation should be based on the original versions of entities, regardless of their attributes \cite{he2017building}. Note that some relation types are antonyms to each other, such as Information--Suggest--Status versus Information--Exclude--Status, Information--Permit--Intervention versus Information--Contra--Intervention. It would be confusing to annotate such relations on entities with a Negation attribute (see Subsection \ref{subsubsec:attr}). For example, the text may state that a negated Information entity suggests a Status entity, which exactly means that the original Information entity excludes the Status entity. In this case, either Information--Suggest--Status or Information--Exclude--Status relation can be justified. As an agreement, we chose to not consider any attributes when annotating relations.

\subsubsection{Medical Attribute} \label{subsubsec:attr}
When annotating an entity in clinical text, we assume by default that the underlying disease (or entity of other types) is currently present on the patient, without explicit uncertainty, conditionality or occasionality, or its status being better or worse. However, the context may deviate from the default. In such case, we would annotate specific attributes on the entity to indicate such deviation. 

Given tokens $x$ and entities $y^{Ent}$, the objective output of attribute extraction is a set $y^{Attr}$, of which each element is a doublet of attribute type and entity:
\begin{align} \label{eq:attribute}
    y^{Attr} = \{ (type^{Attr}_k, entity_k) \mid type^{Attr}_k \in S^{Attr}, \nonumber \\ entity_k \in y^{Ent}, k=1,2,\dots,K \},
\end{align}
where $S^{Attr}$ is the set of attribute types, $K$ is the number of attributes of the entities. Hence, the annotation of a medical attribute should specify its attribute type and associated entity. 

Note that an entity may have multiple attributes. It may seem more straightforward to formulate the attribute extraction as a multi-hot classification task, and use accuracy as the evaluation metric. However, we chose to formulate it by an information retrieval paradigm, just like the entity recognition and relation extraction tasks; accordingly, the annotation and modeling results would also be evaluated using the $F_1$ score (See Subsection \ref{subsec:ann-eval}). By this design, the three tasks would be more unified in annotation, modeling and evaluation.

We formulated ten attribute types. The definitions and examples are listed as follow:
\begin{CJK*}{UTF8}{gbsn}
\begin{enumerate}
    \item \textbf{Negation}: The corresponding entity does not exist, or has not been used; e.g., 否认心脏病~(denies heart disease)
    \item \textbf{Family}: The corresponding entity exists on the patient's family members, rather than the patient; e.g., 患者有偏头痛家族史~(The patients have a family history of migraine)
    \item \textbf{Analysis}: The context is analysis regarding general medical theories and experiences, rather than specified to the current patient. In literature, this attribute may be classified together with Family and described as ``associated with someone other than the patient'' \cite{uzuner2011i2b2}. An Analysis attribute typically appears in clinical analysis, such as prognosis and differential diagnosis. e.g., 该疾病见于老年人，影像学检查可见骨破坏……~(The disease is seen in the elderly, imaging examination shows bone destruction \dots)
    \item \textbf{Uncertainty}: The context expresses significant uncertainty regarding the corresponding entity; e.g., 目前诊断考虑：病毒性肝炎？药物性肝损？~(Current diagnosis considerations: Viral hepatitis? Drug-induced liver damage?)
    \item \textbf{Conditionality}: The underlying symptom only occurs in specific conditions, rather than always exists; e.g., 活动后略感气促~(slightly short of breath after activities)
    \item \textbf{Occasionality}: The underlying symptom occurs occasionally, rather than always exists; e.g., 偶有反酸、嗳气~(occasionally has acid reflux and belching)
    \item \textbf{Better}: The underlying disease or symptom is getting better; e.g., 治疗后疼痛缓解~(The pain got alleviated after treatment)
    \item \textbf{Worse}: The underlying disease or symptom is getting worse; e.g., 两肺多发炎症，对照前片增多~(Multiple inflammations in both lungs, which increased compared with the previous images)
    \item \textbf{History}: The corresponding entity exists in the past; e.g., 患者一周前在家中出现胸闷气促~(The patient had chest tightness and breath shortness at home one week ago)
    \item \textbf{Future}: The corresponding entity can be anticipated in the future; e.g., 疾病进展可出现消化道穿孔等并发症~(Complications such as digestive tract perforation may occur during disease)
\end{enumerate}
\end{CJK*}

Table \ref{tab:attr-applicability} presents the ten medical attribute types and their applicability to entity types. Note that an Abnormal Test Result or Test Result entity should not be attached with a Negation attribute. As aforementioned, these two entity types are the semantically negated versions of each other. In practice, if an Abnormal Test Result can be attached with a Negation attribute, it is equivalent to, and also supposed to be annotated as, a Test Result entity; and vice versa.\footnote{It is necessary to design these two independent entity types (rather than one entity type with a Negation attribute to indicate the difference), because their relations with other entity types are extremely different.} This policy does not apply to other entity types like Self-Reported Abnormality or Test Process; they can be normally negated. 

Following the principle of respecting literalness, an attribute should be annotated only when the local context has explicitly and significantly expressed the corresponding status, which is, by definition, deviated from the default. For example, any diagnoses should be assumed to be uncertain before sufficient medical tests, whereas annotating all these Disease entities with Uncertainty attributes would yield many trivial and redundant annotations. Besides, they are inferred from the global context, and not informative with regard to the local text. Hence, we chose to annotate attributes solely based on the local context. 

\begin{landscape}
\begin{table}
    \caption{Medical attribute types and applicable entity types}
    \centering
    \begin{tabular}{lcccccccc}
        \toprule
                                           & Negation   & Family     & Analysis   & Uncertainty & Conditionality & Occasionality & Better \& Worse & History \& Future \\
        \midrule
        Disease                            & \checkmark & \checkmark & \checkmark & \checkmark  &                &               & \checkmark & \checkmark \\
        Symptom \\ 
        \quad -- Self-Reported Abnormality & \checkmark &            & \checkmark &             & \checkmark     & \checkmark    & \checkmark & \checkmark \\
        \quad -- Abnormal Test Result      &            &            & \checkmark &             & \checkmark     & \checkmark    & \checkmark & \checkmark \\
        Test \\
        \quad -- Test Process              & \checkmark &            & \checkmark &             &                &               &            & \checkmark \\
        \quad -- Test Result               &            &            & \checkmark &             &                &               &            &  \\
        Treatment                          & \checkmark &            & \checkmark &             &                &               &            & \checkmark \\
        Drug                               & \checkmark &            & \checkmark &             &                &               &            & \checkmark \\
        Body                               &            &            & \checkmark &             &                &               &            &  \\
        Personal History                   & \checkmark &            & \checkmark &             &                &               &            &  \\
        Equipment                          &            &            &            &             &                &               &            &  \\
        Department                         &            &            &            &             &                &               &            &  \\
        \bottomrule
    \end{tabular}
    \label{tab:attr-applicability}
\end{table}
\end{landscape}

\subsection{Annotation Tool}
We used the BRAT Rapid Annotation Tool (BRAT) \cite{stenetorp2012brat}. BRAT is designed in particular for structured annotation (including entities, relations and attributes), and provides a native support for Chinese text. It allows flexible customization of  visualizations via the configuration files. In particular, we carefully set up the color scheme for entities, so that different entity types were visually distinctive, and the corresponding semantics could be, to some extent, reflected by the entity colors. For example, the Disease entities were colored in red, whereas the Treatment entities were in green. As reported by the annotators, these configurations provided intuitive feelings of the annotation results, which helped improve the annotation efficiency. 

\begin{figure*}
    \includegraphics[width=\textwidth]{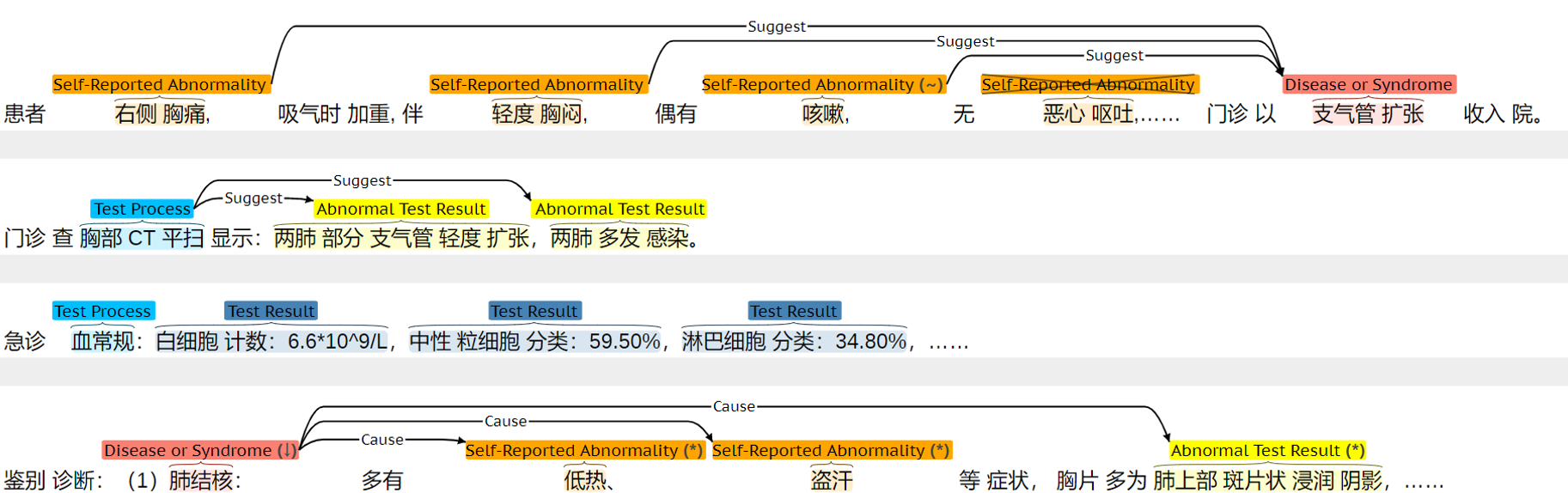}
    \newline \newline
    \footnotesize{\textbf{Translation}: \newline
    \begin{tabular}{p{335pt}}
        The patient had chest pain on the right side, which aggravated during inhalation, accompanied by mild chest tightness, occasionally coughing, no nausea or vomiting, \dots The outpatient department admitted the patient with bronchiectasis. \\
        Outpatient examination of chest CT scan showed: The bronchus of both lungs was slightly dilated, and both lungs were infected with multiple infections. \\
        Emergency blood routine: White blood cell count: 6.6*$10^9$/L, neutrophil classification: 59.50\%, lymphocyte classification: 34.80\%, \dots \\
        Differential diagnosis: (1) Pulmonary tuberculosis: mostly low fever, night sweats with other symptoms, chest images typically show patchy infiltration shadows on the upper lung, \dots 
    \end{tabular}}
    \newline \newline
    \footnotesize{\textbf{Entities}: \newline
    \begin{tabular}{lp{190pt}}
        (Self-Reported Abnormality, 2, 6) & 右侧胸痛~(chest pain on the right side) \\
        (Self-Reported Abnormality, 14, 18) & 轻度胸闷~(mild chest tightness) \\
        (Self-Reported Abnormality, 21, 23) & 咳嗽~(coughing) \\
        (Self-Reported Abnormality, 25, 29) & 恶心呕吐~(nausea or vomiting) \\
        (Disease or Syndrome, 35, 40) & 支气管扩张~(bronchiectasis) \\
        (Test Process, 48, 54) & 胸部CT平扫~(chest CT scan) \\
        \dots & \dots \\
        (Abnormal Test Result, 159, 169) & 肺上部斑片状浸润阴影~(patchy infiltration shadows on the upper lung)
    \end{tabular}}
    \newline \newline
    \footnotesize{\textbf{Relations}: \newline
    \begin{tabular}{l}
        (Information--Suggest--Status, (Self-Reported Abnormality, 2, 6), (Disease or Syndrome, 35, 40)) \\
        (Information--Suggest--Status, (Self-Reported Abnormality, 14, 18), (Disease or Syndrome, 35, 40)) \\
        (Information--Suggest--Status, (Self-Reported Abnormality, 21, 23), (Disease or Syndrome, 35, 40)) \\
        (Information--Suggest--Status, (Test Process, 48, 54), (Abnormal Test Result, 57, 68)) \\
        \dots \\
        (Status--Cause--Information, (Disease or Syndrome, 140, 143), (Abnormal Test Result, 159, 169))
    \end{tabular}}
    \newline \newline
    \footnotesize{\textbf{Attributes}: \newline 
    \begin{tabular}{l}
        (Occasionality, (Self-Reported Abnormality, 21, 23)) \\
        (Negation, (Self-Reported Abnormality, 25, 29)) \\
        (Uncertainty, (Disease or Syndrome, 140, 143)) \\
        (Analysis, (Self-Reported Abnormality, 146, 148)) \\
        (Analysis, (Self-Reported Abnormality, 149, 151)) \\
        (Analysis, (Abnormal Test Result, 159, 169)) 
    \end{tabular}}
    \caption{An example of entity, relation and attribute annotations on BRAT platform}
    \label{fig:brat}
\end{figure*}

Figure \ref{fig:brat} shows an example document along with its medical information annotations on the BRAT platform. The extracted medical entities, relations and attributes are also listed below. This figure demonstrates the input and output formats of our medical information extraction tasks in a fairly straightforward way. 

\begin{CJK*}{UTF8}{gbsn}
\end{CJK*}

\subsection{Annotation Procedure}
Our project team consists of NLP researchers and physicians. The annotation project proceeded in three stages. 
\begin{enumerate}
    \item \textbf{Drafting annotation scheme} (one month): Firstly, the NLP researchers summarized the annotation guidelines in the related literature \cite{uzuner2011i2b2, roberts2009building, he2017building}, and drafted a preliminary annotation scheme. According to this scheme, the BRAT configuration files were written and deployed on the server with about 50 sampled clinical documents. They then tried to annotate these samples and simultaneously revised the scheme and configuration files, ensuring that the annotation system was workable. 
    \item \textbf{Pilot annotation and updating annotation scheme} (one month): Three experienced physicians participated as full-time annotators. They were first illustrated with the project objectives and the draft annotation scheme, and trained on how to annotate on the BRAT platform. Then, each annotator was assigned with 50 sampled full medical records, and attempted to annotate them. Meanwhile, multiple-round meetings were organized between the physicians and NLP researchers to report the issues in annotation practice, and discuss whether and how the annotation scheme should be revised. Actually, numerous issues were reported in the first a few days, so the meetings were scheduled very frequently (almost every half day), and the annotation scheme was correspondingly updated. About two weeks later, the issues became significantly less, and the annotation scheme reached a stable version. 
    \item \textbf{Corpus annotation} (three months): Every week, the three annotators were assigned with about 30 sampled medical records for annotation. They should record the lowly confident cases when annotating the medical records, and report on the weekly meeting for discussion. They might have to revise the annotations according to the discussion results. This stage proceeded until the annotated records could cover representative diseases in the selected medical departments. 
\end{enumerate}

Except for the drafting stage, our annotators were all physicians with clinical experiences; further, they were completely off the clinical work, serving as full-time annotators during the second and third stages. This was costly, but provided a fundamental guarantee of our corpus quality. 

Given the high cost of annotation, we employed two strategies to reduce the burden of annotation labor. The first was the pre-annotation suggested by Campillos \emph{et al.} \cite{campillos2018french}. That is, given a time point, we trained a simple model on the medical records that had been already annotated, and applied it to the unannotated medical records. Thus, the annotators could just revise the incorrect cases, avoiding spending too much time on repetitive work. However, it should also be avoided that the annotators might rely too much on the pre-annotations, so we randomly dropped a proportion of pre-annotations before distributing them to the annotators, which effectively helped the annotators to focus their attention on the annotation work. 

The second strategy was allowing annotating relations between ``entity groups'', as suggested by He \emph{et al.} \cite{he2017building}. A relation annotated between two entity groups means that the relation applies to each pair of entities separately from the two groups. This is suitable for clinical text. For example, it is a common case that multiple medical tests are required to confirm a disease and exclude the differential diagnoses, where a Status--Require--Information relation exists between each pair of disease/diagnosis and medical test. Such group-level relations would be parsed to equivalent entity-level relations before annotation evaluation and modeling.

\subsection{Annotation Evaluation} \label{subsec:ann-eval}
Following previous studies \cite{roberts2009building, he2017building}, we evaluate the corpus annotation consistency by inter-annotator agreement (IAA). We distributed copies of some medical records to different annotators and compared their annotation results. More specifically, we calculated $F_1$ score using one annotator's results $y^\textrm{Ann1}$ as the ground truth, and another annotator's results $y^\textrm{Ann2}$ as the predicted outcome: 
\begin{align}
    P =& \frac{\vert y^\textrm{Ann1} \cap y^\textrm{Ann2} \vert}{\vert y^\textrm{Ann2} \vert}, \\
    R =& \frac{\vert y^\textrm{Ann1} \cap y^\textrm{Ann2} \vert}{\vert y^\textrm{Ann1} \vert}, \\
    F_1 =& \frac{2 \times P \times R}{P + R},
\end{align}

where $y^\textrm{Ann1}$ and $y^\textrm{Ann2}$ can be any one of entity, relation or attribute sets defined as Eq. \eqref{eq:entity}, \eqref{eq:relation} and \eqref{eq:attribute}, respectively. Note that the $F_1$ score remains unchanged if $y^\textrm{Ann1}$ and $y^\textrm{Ann2}$ are swapped.

\subsection{Annotation Results}
The final corpus consists of 1,200 medical records (18,039 documents), which cover the typical diseases in 30 medical departments. 46 medical records (818 documents) were copied and separately distributed to different annotators, and IAAs were calculated by comparing the different annotation versions on same documents. 

In total, the corpus includes 537,230 entities, 152,524 relations and 141,417 attributes. The IAAs of entity, relation and attribute annotations are 94.53\%, 73.73\% and 91.98\%, respectively. Overall, our annotated corpus scale and quality are comparable to, or surpass the corpora by previous studies.

\section{Medical Information Extraction} \label{sec:extraction}
\subsection{Model Structures}
Figure \ref{fig:model} presents the three separate neural network models that we developed to extract the medical entities, relations and attributes from clinical text. The input is a sequence of tokens, or more specifically, Chinese characters \cite{zhang2018chinese}; and the outputs are task-specific, as formulated by Eq. \eqref{eq:entity}, \eqref{eq:relation} and \eqref{eq:attribute}. 

The three models share the same structures in the embedding and encoding layers. An embedder is a table mapping each token to a fixed-dimensional real-valued dense vector, which is also named distributed representations. Current SOTA models for Chinese entity recognition typically enhance the embedder by external lexicons \cite{zhang2018chinese, ma2020simplify}. An encoder can be regarded as a feature extractor, which maps a sequence of representations to another sequence of representations, which are typically fused with contextual information in the original representations. Popular encoder architectures include LSTM \cite{hochreiter1997long} and the Transformer encoder \cite{vaswani2017attention, devlin2019bert}. 

Devlin \emph{et al.} proposed BERT, which was one of the most important recent advances in the NLP field \cite{devlin2019bert}. BERT employs the Transformer encoder, and uses large-scale unannotated corpora for pre-training, via two pre-training objectives, i.e., masked language modeling (MLM) and next sentence prediction (NSP). The pre-trained encoder (along with the embedder), is then jointed with a specific decoder and fine-tuned for downstream task with corresponding annotated corpora. This pre-training and fine-tuning paradigm effectively utilized the semantic patterns hidden in large-scale unannotated corpora, and achieved SOTA results in most NLP tasks. 

The decoders, however, are distinctive between tasks. Their specifications are as follow. 

\paragraph{Entity Recognition} Given the medical entities are flat in our corpus, we transformed the entity recognition to a sequence tagging task via the BIO2 tagging scheme. Specifically, given the input tokens $x = x_1,x_2,\dots,x_T$, each token was labeled with a tag $y_t^{bio}$, which indicates whether the token is the beginning (B) of an entity, inside (I) an entity, or outside (O) all entities (refer Figure \ref{fig:model} as an example). Thus, the objective of sequence tagging is classifying each token to the corresponding tag in the set $S^{bio} = \{ B \mhyphen type^{Ent}, I \mhyphen type^{Ent}, O \mid type^{Ent} \in S^{Ent} \}$, where $S^{Ent}$ is the set of entity types. It could be implemented with a softmax layer: 
\begin{equation}
    \hat{y}_t^{bio} = \softmax \left( W^e h_t + b^e \right),
\end{equation}
which yields a posterior over all potential tags in $S^{bio}$; $h_t \in \mathbb{R}^d$ is the encoder's output representation for $x_t$; $W^e \in \mathbb{R}^{\vert S^{bio} \vert \times d}$ and $b^e \in \mathbb{R}^{\vert S^{bio} \vert}$ are learnable parameters. We additionally used a linear-chain conditional random field (CRF) layer to incorporate the dependence between consecutive tags \cite{collobert2011natural}. The decoded tags could be converted to equivalent entities, which were then fed into the relation and attribute decoders.

\paragraph{Attribute Extraction} We followed the span-based approaches \cite{eberts2019span, zhong2021frustratingly} for both attribute and relation extraction. Specifically, for a given entity $ent = (type^{Ent}, start, end)$, its pooled representation is: 
\begin{equation}
    h_{ent} = f\left(h_{start}, h_{start+1}, \dots, h_{end-1} \right),
\end{equation}
where $f(\cdot)$ is a function aggregating the representations over the entity span; we chose the max-pooling following Eberts and Ulges \cite{eberts2019span}. The pooled representation was then used to assign the corresponding entity with attribute types in the set $S^{Attr} \cup \{ None \}$, where $None$ is a special type indicating nonexistence of any attributes. Since an entity may have multiple attributes, the classifier should be a multi-hot sigmoid layer: 
\begin{equation}
    \hat{y}_{ent}^{Attr} = \sigmiod \left( W^a h_{ent} + b^a \right),
\end{equation}
which yields a probability vector of all potential attribute types in $S^{Attr} \cup \{ None \}$; $W^a \in \mathbb{R}^{(\vert S^{Attr} \vert + 1) \times d}$ and $b^a \in \mathbb{R}^{\vert S^{Attr} \vert + 1}$ are learnable parameters. Any probability in $\hat{y}_{ent}^{Attr}$ higher than a threshold $\alpha$ indicates the existence of corresponding attribute, unless the probability of $None$ is higher than $\alpha$. 

\paragraph{Relation Extraction} Given two candidate entities $head$ and $tail$, their pooled representations $h_{head}$ and $h_{tail}$ were concatenated and then fed into a softmax layer: 
\begin{equation}
    \hat{y}_{head, tail}^{Rel} = \softmax \left( W^r \left[ h_{head} ; h_{tail} \right] + b^r \right),
\end{equation}
which yields a posterior over all potential relation types $S^{Rel} \cup \{ None \}$, where $None$ is a special type indicating nonexistence of any relations; $W^r \in \mathbb{R}^{(\vert S^{Rel} \vert + 1) \times 2d}$ and $b^r \in \mathbb{R}^{\vert S^{Rel} \vert + 1}$ are learnable parameters. 

The three models have the same structures of embedder and encoder, so they can share the corresponding weights and fit the training data by end-to-end joint training. However, we chose to follow a pipeline approach as suggested by Zhong and Chen \cite{zhong2021frustratingly}; that is, we separately trained the three models, and used the outputs of the entity recognition model as the inputs to the relation and attribute extraction models. According to our experiments, it is more robust to train pipeline models, which converge easily and quickly, and typically achieve performance comparable to a joint model. 

\begin{landscape}
\begin{CJK*}{UTF8}{gbsn}
\begin{figure}
    \centering
    \begin{tikzpicture}[scale=1.0, transform shape, 
                        layer/.style   = {rectangle, thick, rounded corners=2pt, minimum width=380pt, minimum height=30pt, font=\footnotesize}, 
                        charnode/.style = {rectangle, thick, rounded corners=2pt, minimum size=16pt, font=\scriptsize}, 
                        recnode/.style = {rectangle, thick, rounded corners=2pt, minimum width=16pt, minimum height=8pt, font=\scriptsize}, 
                        cirnode/.style = {circle, thick, minimum size=16pt},
                        recbox/.style  = {rectangle, densely dotted, thick, rounded corners=2pt, minimum width=10pt, minimum height=35pt},
                        invnode/.style = {circle}, 
                        red7020/.style    = {draw=casred!70, fill=casred!20},
                        yellow7020/.style = {draw=casyellow!70, fill=casyellow!20}, 
                        green7020/.style  = {draw=casgreen!70, fill=casgreen!20}, 
                        blue7020/.style   = {draw=casblue!70, fill=casblue!20}, 
                        purple7020/.style = {draw=caspurple!70, fill=caspurple!20}, 
                        gray7020/.style   = {draw=gray!70, fill=gray!20}]
        \node[layer, gray7020] at (0pt, 35pt) {LSTM, BERT, etc.};
        
        \node[align=center, font=\footnotesize, text width=80pt] at (-225pt, 0pt) {\textit{Tokens}}; 
        \node[align=center, font=\footnotesize, text width=80pt] at (-225pt, 35pt) {Embedder \\ \& Encoder}; 
        \node[align=center, font=\footnotesize, text width=80pt] at (-225pt, 70pt) {\textit{Representations}}; 
        \node[align=center, font=\footnotesize, text width=80pt] at (-225pt, 95pt) {Ent. Decoder \\ (CRF Layer)};
        \node[align=center, font=\footnotesize, text width=80pt] at (-225pt, 150pt) {Attr. Decoder \\ (Classifier)};
        \node[align=center, font=\footnotesize, text width=80pt] at (-225pt, 205pt) {Rel. Decoder \\ (Classifier)};
        
        \foreach \x \c  in {-180/患, -160/者, -140/有, -120/低, -100/热, -80/、, -60/咳, -40/嗽, -20/症, 0/状, 20/，, 40/考, 60/虑, 80/肺, 100/结, 120/核, 140/可, 160/能, 180/。}{
            \node[charnode, gray7020] (char)  at (\x pt, 0pt) {\c};
            \node[invnode]   (inv1)  at (\x pt, 20pt) {};
            \draw[-stealth] (char.north) -- (inv1.center);
        }
        \foreach \x in {-180, -160, -140, -80, -20, 0, 20, 40, 60, 140, 160, 180}{
            \node[invnode]  (inv2)  at (\x pt, 50pt) {};
            \node[recnode, gray7020] (repr) at (\x pt, 70pt) {};
            \node[recnode, gray7020] (ent) at (\x pt, 95pt) {O};
            \draw[-stealth] (inv2.center) -- (repr.south);
            \draw[-stealth] (repr.north) -- (ent.south);
        }
        \node[invnode]  (inv2)  at (-120pt, 50pt) {};
        \node[recnode, red7020] (repr) at (-120pt, 70pt) {};
        \node[recnode, blue7020] (ent) at (-120pt, 95pt) {B-S};
        \draw[-stealth] (inv2.center) -- (repr.south);
        \draw[-stealth] (repr.north) -- (ent.south);
        
        \node[invnode]  (inv3)  at (-120pt, 110pt) {};
        \draw[-, densely dashed] (repr.north) to [bend left] (inv3.center) node [left, font=\scriptsize] {Pooling};
        
        \node[invnode]  (inv2)  at (-100pt, 50pt) {};
        \node[recnode, red7020] (repr) at (-100pt, 70pt) {};
        \node[recnode, blue7020] (ent) at (-100pt, 95pt) {I-S};
        \draw[-stealth] (inv2.center) -- (repr.south);
        \draw[-stealth] (repr.north) -- (ent.south);
        
        \node[invnode] (inv4)  at (-100pt, 110pt) {};
        \draw[-, densely dashed] (repr.north) to [bend left] (inv4.center);
        \draw[-] (inv3.center) -- (inv4.center);
        
        \node[invnode] (inv5) at (-110pt, 110pt) {};
        \node[recnode, red7020] (repr) at (-110pt, 125pt) {};
        \draw[-stealth] (inv5.center) -- (repr.south);
        
        \node[recnode, green7020] (attr) at (-110pt, 150pt) {None};
        \draw[-stealth] (repr.north) -- (attr.south);
        
        \node[invnode] (inv6) at (-110pt, 165pt) {};
        \draw[-, densely dashed] (repr.north) to [bend left] (inv6.center);
        
        \node[invnode]  (inv2)  at (-60pt, 50pt) {};
        \node[recnode, red7020] (repr) at (-60pt, 70pt) {};
        \node[recnode, blue7020] (ent) at (-60pt, 95pt) {B-S};
        \draw[-stealth] (inv2.center) -- (repr.south);
        \draw[-stealth] (repr.north) -- (ent.south);
        
        \node[invnode]  (inv3)  at (-60pt, 110pt) {};
        \draw[-, densely dashed] (repr.north) to [bend left] (inv3.center) node [left, font=\scriptsize] {Pooling};
        
        \node[invnode]  (inv2)  at (-40pt, 50pt) {};
        \node[recnode, red7020] (repr) at (-40pt, 70pt) {};
        \node[recnode, blue7020] (ent) at (-40pt, 95pt) {I-S};
        \draw[-stealth] (inv2.center) -- (repr.south);
        \draw[-stealth] (repr.north) -- (ent.south);
        
        \node[invnode] (inv4)  at (-40pt, 110pt) {};
        \draw[-, densely dashed] (repr.north) to [bend left] (inv4.center);
        \draw[-] (inv3.center) -- (inv4.center);
        
        \node[invnode] (inv5) at (-50pt, 110pt) {};
        \node[recnode, red7020] (repr) at (-50pt, 125pt) {};
        \draw[-stealth] (inv5.center) -- (repr.south);
        
        \node[recnode, green7020] (attr) at (-50pt, 150pt) {None};
        \draw[-stealth] (repr.north) -- (attr.south);
        
        \node[invnode] (inv7) at (-50pt, 165pt) {};
        \draw[-, densely dashed] (repr.north) to [bend left] (inv7.center);
        
        \draw[-] (inv6.center) -- (inv7.center) node [midway, below, font=\scriptsize] {Concat.};
        \node[invnode] (inv8) at (-80pt, 165pt) {};
        \node[recnode] (repr) at (-80pt, 180pt) {};
        \node[recnode, red7020] at (-88pt, 180pt) {};
        \node[recnode, red7020] at (-72pt, 180pt) {};
        \draw[-stealth] (inv8.center) -- (repr.south);
        
        \node[recnode, yellow7020] (rel) at (-80pt, 205pt) {None};
        \draw[-stealth] (repr.north) -- (rel.south);
        
        \node[invnode]  (inv2)  at (80pt, 50pt) {};
        \node[recnode, red7020] (repr) at (80pt, 70pt) {};
        \node[recnode, blue7020] (ent) at (80pt, 95pt) {B-D};
        \draw[-stealth] (inv2.center) -- (repr.south);
        \draw[-stealth] (repr.north) -- (ent.south);
        
        \node[invnode]  (inv3)  at (80pt, 110pt) {};
        \draw[-, densely dashed] (repr.north) to [bend left] (inv3.center) node [left, font=\scriptsize] {Pooling};
        
        \node[invnode]  (inv2)  at (100pt, 50pt) {};
        \node[recnode, red7020] (repr) at (100pt, 70pt) {};
        \node[recnode, blue7020] (ent) at (100pt, 95pt) {I-D};
        \draw[-stealth] (inv2.center) -- (repr.south);
        \draw[-stealth] (repr.north) -- (ent.south);
        
        \node[invnode] (inv5)  at (100pt, 110pt) {};
        \draw[-, densely dashed] (repr.north) to [bend left] (inv5.center);
        
        \node[invnode]  (inv2)  at (120pt, 50pt) {};
        \node[recnode, red7020] (repr) at (120pt, 70pt) {};
        \node[recnode, blue7020] (ent) at (120pt, 95pt) {I-D};
        \draw[-stealth] (inv2.center) -- (repr.south);
        \draw[-stealth] (repr.north) -- (ent.south);
        
        \node[invnode] (inv4)  at (120pt, 110pt) {};
        \draw[-, densely dashed] (repr.north) to [bend left] (inv4.center);
        
        \draw[-] (inv3.center) -- (inv4.center);
        \node[recnode, red7020] (repr) at (100pt, 125pt) {};
        \draw[-stealth] (inv5.center) -- (repr.south);
        
        \node[recnode, green7020] (attr) at (100pt, 150pt) {Uncertainty};
        \draw[-stealth] (repr.north) -- (attr.south);
        
        \node[invnode] (inv8) at (100pt, 165pt) {};
        \draw[-, densely dashed] (repr.north) to [bend left] (inv8.center);
        
        \draw[-] (inv7.center) -- (inv8.center) node [midway, below, font=\scriptsize] {Concat.};
        \node[invnode] (inv9) at (25pt, 165pt) {};
        \node[recnode] (repr) at (25pt, 180pt) {};
        \node[recnode, red7020] at (17pt, 180pt) {};
        \node[recnode, red7020] at (33pt, 180pt) {};
        \draw[-stealth] (inv9.center) -- (repr.south);
        
        \node[recnode, yellow7020] (rel) at (25pt, 205pt) {Suggest};
        \draw[-stealth] (repr.north) -- (rel.south);
    \end{tikzpicture}
    \caption{The model structures for entity, relation and attribute extraction. ``B-S'', ``I-S'', ``B-D'', ``I-D'' are short for ``B-Symptom'', ``I-Symptom'', ``B-Disease'', ``I-Disease'', respectively.}
    \label{fig:model}
\end{figure}
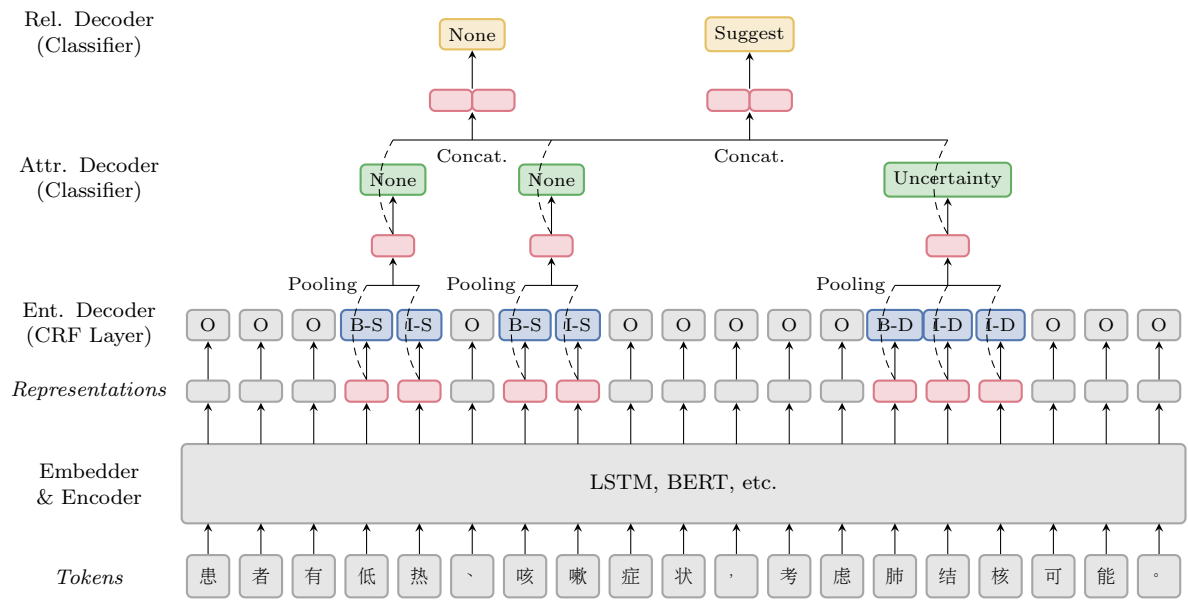
\end{CJK*}
\end{landscape}

\subsection{Experimental Settings}
We performed experiments on the released corpus, HwaMei-500. It was split into training, development and testing sets, which comprise 300, 100 and 100 annotated medical records, respectively. The development set was used for selecting hyper-parameters, and the testing set was used to calculate the evaluation metrics, i.e., precision rate, recall rate and $F_1$ score. Consistent with previous IAA evaluation, a tuple (entity, relation or attribute) is considered correct only if all its components are correct. 

We used either a character-based LSTM or a pre-trained language model as the embedder/encoder structure. The character-based LSTM was trained from scratch, and we optionally enhanced it by an additional bigram or SimpleLexicon embedder \cite{ma2020simplify}. The pre-trained language models are BERT \cite{devlin2019bert}; BERT-wwm and RoBERTa-wwm, trained with whole word masking \cite{cui2019pretraining}; MacBERT, trained by MLM as correction \cite{cui2020revisiting}; ERNIEv1, trained with entity- and phrase-level masking guided by external knowledge \cite{sun2019ernie}. They are pre-trained on Chinese corpora. These specifications could be regarded as representative and strong baselines for Chinese information extraction tasks. 

The character embedding dimension is 200, and the LSTM has two layers of hidden size 600. The character embeddings were randomly initialized by a uniform distribution between the range $\left[-\sqrt{\frac{3}{dim}}, \sqrt{\frac{3}{dim}}\right]$, where $dim$ was the embedding dimension \cite{he2015delving}; other model weights were initialized by the Xavier initialization \cite{glorot2010understanding}, except the pre-trained ones. Dropout was applied between neural layers with a rate of 0.5 \cite{srivastava2014dropout}. All models were trained by the AdamW optimizer \cite{loshchilov2018decoupled} with a gradient clipping at L2-norm of 5.0 \cite{pascanu2013difficulty}. Character-based LSTM models were trained with learning rate 1e-3, batch size 48 for 50 epochs; while pre-trained models were trained with learning rate 2e-5, batch size 48 for 25 epochs (linear warmup in the first 20\% steps followed by a linear decay). Early stopping was applied by monitoring the $F_1$ scores on the development set.

\subsection{Benchmarking Results}
Table \ref{tab:performance} presents the performance of our medical information extraction models on HwaMei-500. We report the average and standard deviation across five runs, for each evaluation on the testing set. The best testing $F_1$ scores are 92.23\%, 61.46\% and 87.51\%, for entity recognition, relation extraction and attribute extraction, respectively. These numbers are not far away from the corresponding IAAs, which suggests that these models may collaboratively serve as a workable medical information extraction system.  

A simple character-based LSTM can already achieve relatively high $F_1$ scores. In literature, it is thought of as a robust boosting approach to add bigram or lexicon-enhanced embedder for Chinese entity recognition \cite{zhang2018chinese, ma2020simplify}. However, our experiments show that both of them are generally ineffective. It is probably because that the lexicon used here is for common text, rather than medical text. Further experiments are required to verify this point. 

The pre-trained language models achieve very similar performance, but all significantly outperform the character-based LSTM. Note that these language models are pre-trained on common text datasets (Chinese Wikipedia, web news, etc.), which have limited medicine-related contents. Hence, the results suggest that the universal linguistic patterns are also quite useful for medical information extraction. Additionally, a LSTM layer above the pre-trained models can further consistently improve the results, especially for the relation extraction. ``RoBERTa-wwm + LSTM'' seems to be the best specification. This is plausibly because that RoBERTa focuses on the token-level task (i.e., MLM) and removes the sequence-level objective (i.e., NSP) \cite{liu2019roberta}; thus RoBERTa is particularly effective for within-sequence downstream tasks, e.g., information extraction. 

\begin{landscape}
\begin{table}
    \caption{Performance of medical information extraction models on HwaMei-500}
    \centering
    \begin{tabular}{lccccccccc}
        \toprule
         & \multicolumn{3}{c}{Entity} & \multicolumn{3}{c}{Relation} & \multicolumn{3}{c}{Attribute} \\
        \cmidrule(lr){2-4} \cmidrule(lr){5-7} \cmidrule(lr){8-10} 
         & Prec. & Rec. & F1 & Prec. & Rec. & F1 & Prec. & Rec. & F1 \\
        \midrule
        Char-based LSTM       & 90.94$_{\pm 0.11}$ & 91.09$_{\pm 0.08}$ & 91.02$_{\pm 0.04}$ & 50.64$_{\pm 2.25}$ & 59.10$_{\pm 1.04}$ & 54.50$_{\pm 0.86}$ & 86.52$_{\pm 0.22}$ & 82.58$_{\pm 0.41}$ & 84.50$_{\pm 0.17}$ \\
        \quad + Bigram        & 90.14$_{\pm 0.22}$ & 91.63$_{\pm 0.11}$ & 90.88$_{\pm 0.15}$ & 50.84$_{\pm 2.06}$ & 60.31$_{\pm 0.99}$ & 55.14$_{\pm 0.97}$ & 85.72$_{\pm 0.37}$ & 83.26$_{\pm 0.23}$ & 84.47$_{\pm 0.08}$ \\
        \quad + SimpleLexicon & 90.07$_{\pm 0.17}$ & 91.77$_{\pm 0.14}$ & 90.91$_{\pm 0.15}$ & 46.52$_{\pm 0.92}$ & 62.48$_{\pm 0.58}$ & 53.33$_{\pm 0.57}$ & 85.23$_{\pm 0.44}$ & 83.53$_{\pm 0.40}$ & 84.37$_{\pm 0.07}$ \\
        \midrule
        BERT                  & 91.51$_{\pm 0.43}$ & 92.49$_{\pm 0.32}$ & 92.00$_{\pm 0.37}$ & 46.94$_{\pm 1.84}$ & 66.81$_{\pm 2.15}$ & 55.10$_{\pm 1.02}$ & 86.22$_{\pm 0.49}$ & 86.90$_{\pm 0.21}$ & 86.56$_{\pm 0.16}$ \\
        \quad + LSTM          & 91.47$_{\pm 0.45}$ & 92.63$_{\pm 0.27}$ & 92.05$_{\pm 0.36}$ & 57.53$_{\pm 1.66}$ & 65.54$_{\pm 0.44}$ & 61.26$_{\pm 0.88}$ & 86.66$_{\pm 0.21}$ & 87.39$_{\pm 0.14}$ & 87.02$_{\pm 0.14}$ \\
        \midrule
        BERT-wwm              & 91.48$_{\pm 0.54}$ & 92.61$_{\pm 0.29}$ & 92.04$_{\pm 0.42}$ & 41.70$_{\pm 0.82}$ & 69.20$_{\pm 1.01}$ & 52.04$_{\pm 0.79}$ & 86.69$_{\pm 0.26}$ & 87.08$_{\pm 0.16}$ & 86.88$_{\pm 0.12}$ \\
        \quad + LSTM          & 91.57$_{\pm 0.33}$ & 92.78$_{\pm 0.28}$ & 92.17$_{\pm 0.30}$ & 55.72$_{\pm 1.90}$ & 66.92$_{\pm 0.55}$ & 60.79$_{\pm 1.03}$ & 86.49$_{\pm 0.25}$ & 87.42$_{\pm 0.25}$ & 86.95$_{\pm 0.12}$ \\
        \midrule
        RoBERTa-wwm           & 91.63$_{\pm 0.52}$ & 92.49$_{\pm 0.35}$ & 92.06$_{\pm 0.43}$ & 44.75$_{\pm 1.22}$ & 68.80$_{\pm 1.77}$ & 54.20$_{\pm 0.71}$ & 86.53$_{\pm 0.23}$ & 87.72$_{\pm 0.14}$ & 87.12$_{\pm 0.17}$ \\
        \quad + LSTM          & 91.76$_{\pm 0.33}$ & 92.71$_{\pm 0.19}$ & \textbf{92.23}$_{\pm 0.26}$ & 57.53$_{\pm 1.34}$ & 65.98$_{\pm 0.33}$ & \textbf{61.46}$_{\pm 0.79}$ & 86.87$_{\pm 0.24}$ & 88.16$_{\pm 0.13}$ & \textbf{87.51}$_{\pm 0.10}$ \\
        \midrule
        MacBERT               & 91.66$_{\pm 0.35}$ & 92.48$_{\pm 0.18}$ & 92.07$_{\pm 0.26}$ & 44.36$_{\pm 1.43}$ & 68.67$_{\pm 1.32}$ & 53.88$_{\pm 1.01}$ & 86.56$_{\pm 0.28}$ & 87.55$_{\pm 0.28}$ & 87.05$_{\pm 0.13}$ \\
        \quad + LSTM          & 91.72$_{\pm 0.34}$ & 92.69$_{\pm 0.19}$ & 92.20$_{\pm 0.26}$ & 56.21$_{\pm 2.12}$ & 67.24$_{\pm 0.56}$ & 61.20$_{\pm 1.09}$ & 86.77$_{\pm 0.18}$ & 87.60$_{\pm 0.10}$ & 87.18$_{\pm 0.06}$ \\
        \midrule
        ERNIEv1               & 91.23$_{\pm 0.61}$ & 91.96$_{\pm 0.46}$ & 91.59$_{\pm 0.53}$ & 43.05$_{\pm 2.92}$ & 67.07$_{\pm 2.45}$ & 52.33$_{\pm 1.61}$ & 86.13$_{\pm 0.36}$ & 87.24$_{\pm 0.08}$ & 86.68$_{\pm 0.16}$ \\
        \quad + LSTM          & 91.59$_{\pm 0.45}$ & 92.34$_{\pm 0.33}$ & 91.97$_{\pm 0.38}$ & 57.07$_{\pm 1.43}$ & 64.64$_{\pm 0.45}$ & 60.61$_{\pm 0.85}$ & 86.78$_{\pm 0.34}$ & 87.05$_{\pm 0.30}$ & 86.92$_{\pm 0.07}$ \\
        \midrule
        IAA                   &  &  & 94.53 &  &  & 73.73 &  &  & 91.98 \\
        \bottomrule
    \end{tabular}
    \newline
    \footnotesize{The reported metrics are the average and standard deviation across five runs with different random seeds. All results are evaluated on the testing set. The best $F_1$ score on each task is highlighted in bold.}
    \label{tab:performance}
\end{table}
\end{landscape}

\subsection{The Effect of More Data}
\begin{figure}[!ht]
    \centering
    \begin{subfigure}{0.45\textwidth}
    \centering
    \includegraphics[width=\textwidth]{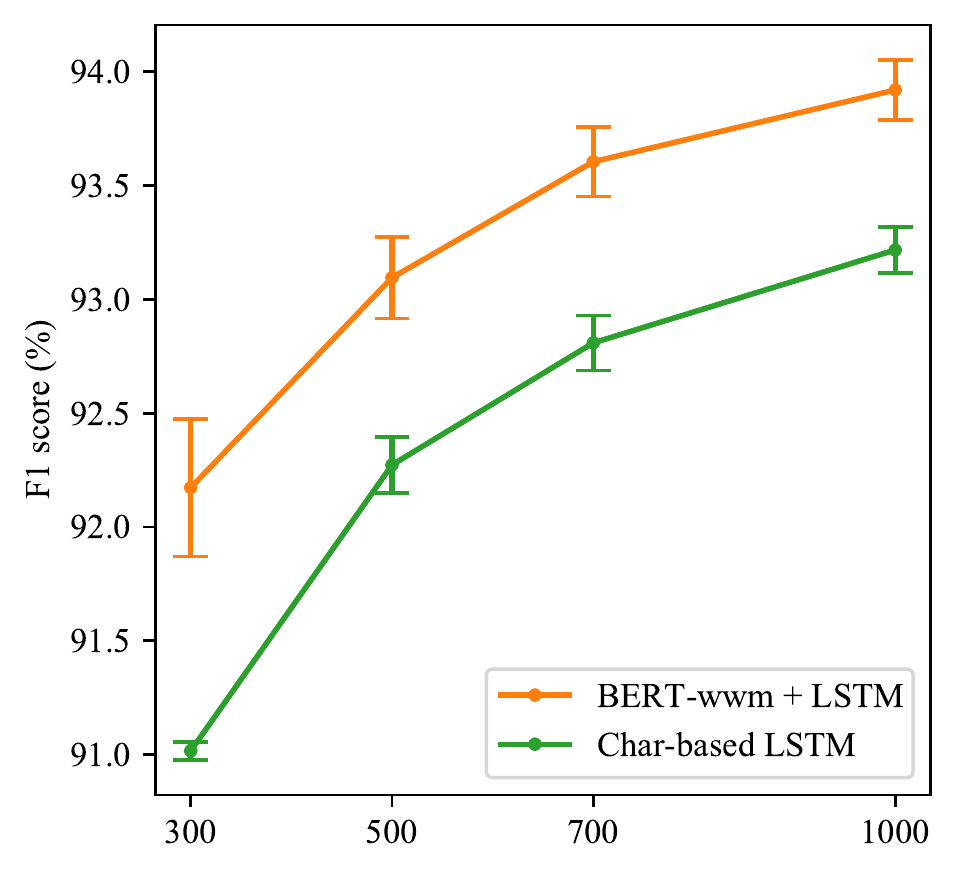}
    \caption{Entity recognition}
    \end{subfigure}
    \begin{subfigure}{0.45\textwidth}
    \centering
    \includegraphics[width=\textwidth]{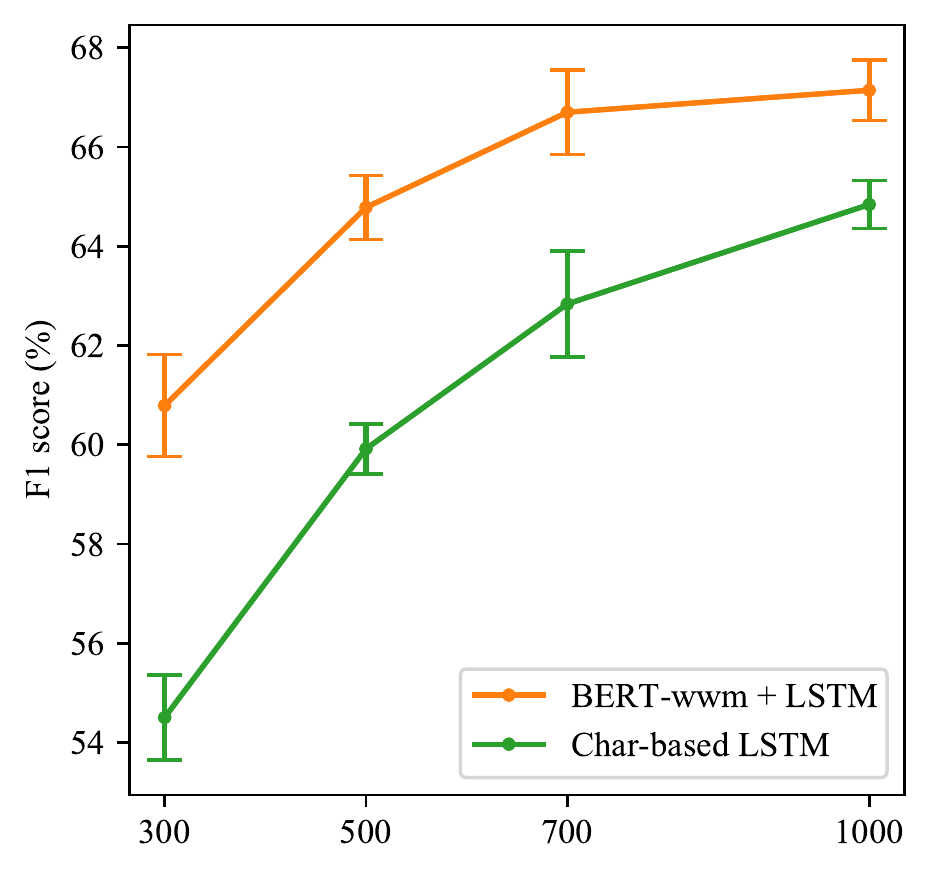}
    \caption{Relation extraction}
    \end{subfigure}
    \begin{subfigure}{0.45\textwidth}
    \centering
    \includegraphics[width=\textwidth]{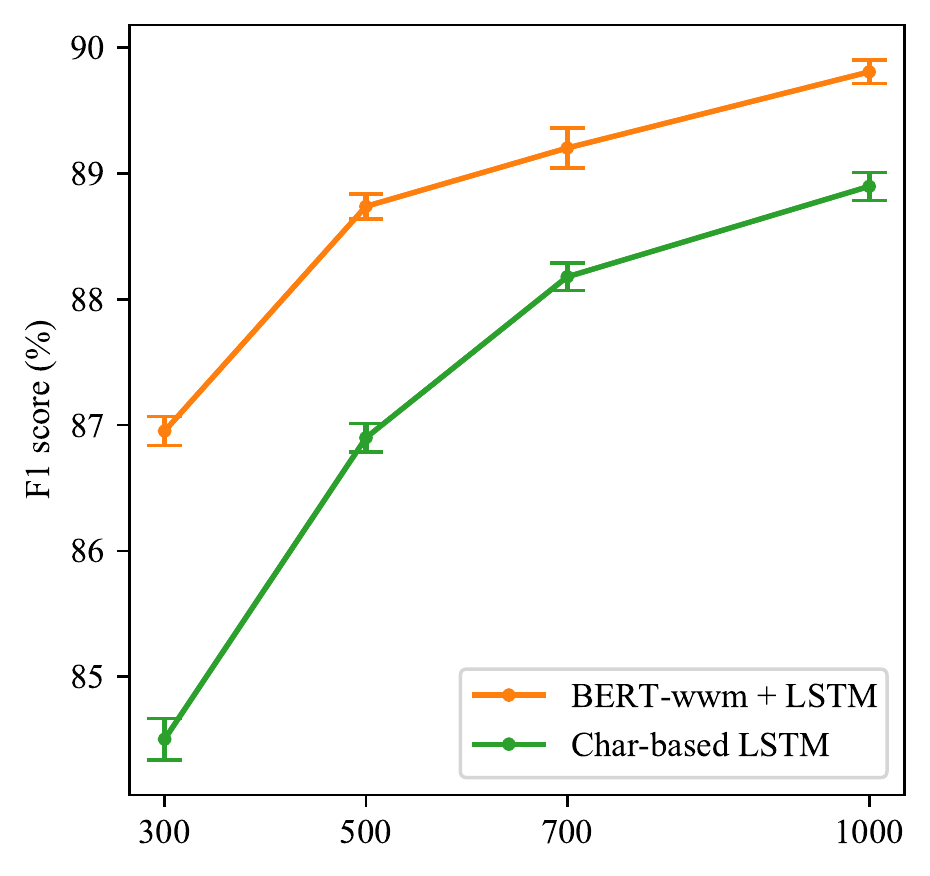}
    \caption{Attribute extraction}
    \end{subfigure}
    \caption{Medical information extraction performance increases with more training data. The reported $F_1$ scores are the average across five runs with different random seeds, and the error bars represent corresponding standard deviation. All the experiments use the same development and testing sets as HwaMei-500.}
    \label{fig:f1-scores}
\end{figure}

Figure \ref{fig:f1-scores} presents how the model performance scales up with the corpus size. For a fair comparison, all the experiments use the same development and testing sets as HwaMei-500. Overall, all three tasks consistently and progressively benefit from more training data, regardless of the neural structure used. In the case of ``BERT-wwm + LSTM'', the additional 700 annotated medical records can boost the $F_1$ score by 1.75 percentages (92.17\% to 93.92\%) for entity recognition, 6.36 percentages (60.79\% to 67.15\%) for relation extraction, and 3.86 percentages (85.95\% to 89.81\%) for attribute extraction. Apparently, the improvements are much larger than the performance gains by model structures (i.e., pre-trained Transformer vs. LSTM). This is consistent to the common knowledge that deep learning models are data-driven \cite{lecun2015deep,hirschberg2015advances}.

Note that the $F_1$ scores approach the corresponding IAAs when more training data are leveraged. Although the marginal gains present a diminishing trend, the effect of more training data does not saturate. This displays a promising path towards a medical information extraction system that performs comparable to, or even better than human annotation. In addition, the trends in Figure \ref{fig:f1-scores} also provide the practitioners a clear guidance to what extent they should extend the corpus, if necessary, to achieve their desired performance. 

\section{Conclusion} \label{sec:conclusion}
This article presents a clinical-text-oriented medical information engineering framework, consisting of three tasks: medical entity recognition, relation extraction and attribute extraction. We in detail demonstrate the whole workflow, from EMR data collection through performance evaluation. We have released our annotation scheme, annotated corpus (HwaMei-500) and code. We believe that these materials are insightful and useful for either researchers or practitioners, and will facilitate future research. 

Our annotation scheme is comprehensive and compatible between tasks; the corpus (HwaMei-1200) includes 1,200 full medical records manually annotated by experienced physicians. To the best of our knowledge, this is the first Chinese medical information extraction corpus of such scale and quality. Our experiments show that the model performance can approach that of human annotation, if more training data are available. 

Future work may introduce the entity normalization task, following the studies in the biomedical domain \cite{dougan2014ncbi, li2016biocreative}. In addition, we will leverage priori medical knowledge in both pre-training and fine-tuning phases \cite{sun2019ernie}, to further boost the information extraction performance.

\section*{Acknowledgment}
This work was supported by Ningbo Science and Technology Service Industry Demonstration Project, China (2020F041). We highly appreciate the constructive comments on the annotation scheme by Xiaofen Zhao. We sincerely thank Xin Wei for his efforts in EMR data cleaning and parsing.

\bibliography{references}

\end{document}